\documentclass[journal]{IEEEtran}
\usepackage{times}
\usepackage{epsfig}
\usepackage{graphicx}
\usepackage{amsmath}
\usepackage{amssymb}
\usepackage{color}
\usepackage{booktabs}
\usepackage{subcaption}
\usepackage{bm}
\usepackage{multirow}
\usepackage{cite}
\usepackage{latexsym}

\begin{document}
\title{TTPP: Temporal Transformer with Progressive Prediction for Efficient Action Anticipation}

\author{Wen~Wang, Xiaojiang~Peng, Yanzhou~Su, Yu~Qiao, Jian~Cheng
\thanks{Wen Wang, Yanzhou Su and Jian Cheng are with the School of Information and Communication Engineering, University of Electronic Science and Technology of China, Chengdu, Sichuan, China, 611731.}
\thanks{Xiaojiang Peng and Yu Qiao are with ShenZhen Key Lab of Computer Vision and Pattern Recognition, SIAT-SenseTime Joint Lab, Shenzhen Institutes of Advanced Technology, Chinese Academy of Sciences; SIAT Branch, Shenzhen Institute of Artificial Intelligence and Robotics for Society.}
\thanks{This work was done when Wen Wang was intern at Shenzhen Institutes of Advanced Technology, Chinese Academy of Sciences.}
\thanks{Corresponding author: Xiaojiang Peng (xj.peng@siat.ac.cn)}}

\newcommand{\etal}{\textit{et al}. }
\newcommand{\ie}{\textit{i}.\textit{e}. }
\newcommand{\eg}{\textit{e}.\textit{g}. }

\markboth{Journal of \LaTeX\ Class Files,~Vol.~14, No.~8, August~2015}%
{Shell \MakeLowercase{\textit{et al.}}: Bare Demo of IEEEtran.cls for IEEE Journals}

\maketitle
\begin{abstract}
Video action anticipation aims to predict future action categories from observed frames. Current state-of-the-art approaches mainly resort to recurrent neural networks to encode history information into hidden states, and predict future actions from the hidden representations. It is well known that the recurrent pipeline is inefficient in capturing long-term information which may limit its performance in predication task. To address this problem, this paper proposes a simple yet efficient Temporal Transformer with Progressive Prediction (TTPP) framework, which repurposes a Transformer-style architecture to aggregate observed features, and then leverages a light-weight network to progressively predict future features and actions. Specifically, predicted features along with predicted probabilities are accumulated into the inputs of subsequent prediction. We evaluate our approach on three action datasets, namely TVSeries, THUMOS-14, and TV-Human-Interaction. Additionally we also conduct a comprehensive study for several popular aggregation and prediction strategies. Extensive results show that TTPP not only outperforms the state-of-the-art methods but also more efficient.
\end{abstract}

\section{Introduction}
Human action anticipation, also aka early action prediction, aiming to predict future unseen actions, is one of the main topics in video understanding with wide applications in security, visual surveillance and human-computer interaction, etc. In contrast to the well-studied action recognition, which infers the action label after observing the entire action execution, action anticipation is to early predict human actions without observing the future action execution. It is a very challenging task because the input videos are temporally incomplete with wide variety of irrelevant background, and decisions must be made based on such incomplete action executions. In all, action anticipation needs to overcome all the difficulties of action recognition and capture sufficient historical and contextual information to make future predictions in untrimmed video streams.

\begin{figure}
\begin{center}
\includegraphics[width=8.7cm]{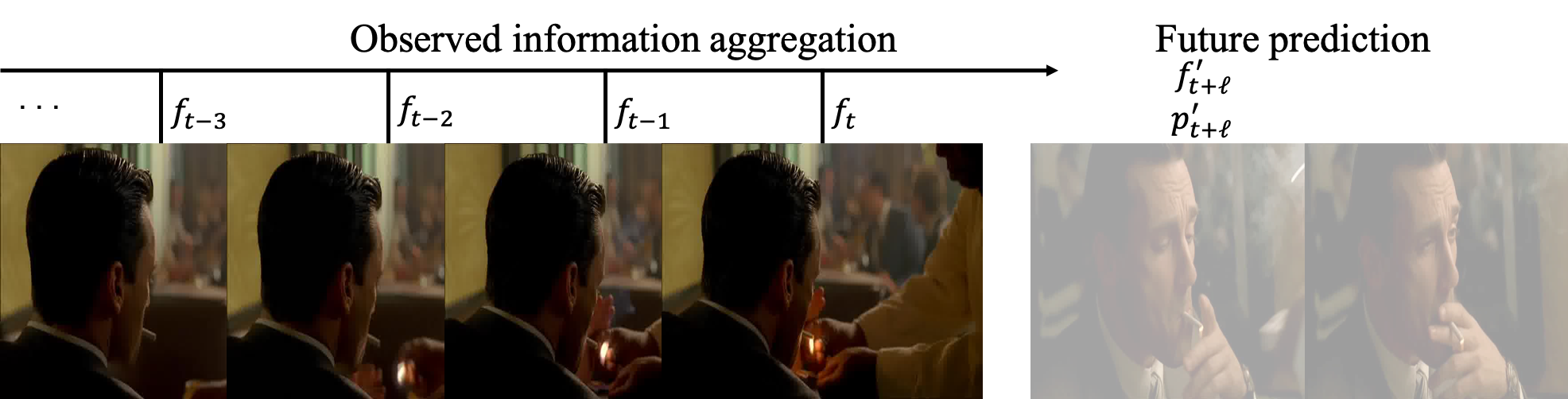}
\end{center}
   \caption{The summarized generic flowchart for action anticipation, which mainly consists of observed information aggregation and future prediction.}
\label{fig_1}
\end{figure}

Generally, most of the action anticipation approaches can be divided into two key phases, namely observed information aggregation and future prediction, as shown in Figure \ref{fig_1}. Early works of action anticipation focus on trimmed action videos, and mainly make efforts on extracting discriminative features from partial videos, \ie observed information aggregation, for early action prediction \cite{Aliakbarian2017Encouraging, Kong2014A, 8099873, Ryoo2012Human, Wang2015Towards}.
In deep learning era, recent works turn to predict future actions in practical untrimmed video streams \cite{Furnari_2019_ICCV, DBLP:conf/bmvc/GaoYN17a, Wang2019Delving, Xu_2019_ICCV}, and mainly repurpose sequential models from the natural language processing (NLP) domain, like long short-term memory (LSTM)~\cite{Graves1997Long} and gated recurrent neural networks~\cite{DBLP:journals/corr/ChungGCB14}. For instance, Gao \etal \cite{DBLP:conf/bmvc/GaoYN17a} propose a Reinforced Encoder-Decoder network, which utilizes an encoder-decoder LSTM network~\cite{Luong2015Effective, Wu2016Google} to aggregate historical features and predict future features or actions.
Xu \etal \cite{Xu_2019_ICCV} propose a LSTM-based temporal recurrent network to predict future features for both online action detection and action anticipation. Though the encoder-decoder recurrent networks can be easily transformed from NLP domain to temporal action anticipation, their inherent sequential nature precludes parallelization within training examples and limits the memory power for longer sequence length \cite{DBLP:journals/corr/VaswaniSPUJGKP17}. Moreover, they are known to have limited improvements in other action understanding tasks \cite{Donahue_2015_CVPR, DBLP:journals/corr/WuWJYX15}.

In this paper, we address the two issues of action anticipation via a simple yet efficient \textit{Temporal Transformer with Progressive Prediction} (TTPP) framework. TTPP repurposes a Transformer-style module to aggregate observed information and leverages a light-weight network to progressively predict future features and actions. Specifically, TTPP contains a Temporal Transformer Module (TTM) and an elaborately-designed Progressive Prediction Module (PPM). Given historical and current features, the TTM aggregates the historical features based solely on self-attention mechanisms with the current feature as query, which is inspired by the Transformer in machine translation \cite{DBLP:journals/corr/VaswaniSPUJGKP17}. The aggregated historical feature along with the current feature are then fed into the PPM. The PPM is comprised of an initial prediction block and a shared-parameter progressive prediction block, each of which is built with two fully-connected (FC) layers, a ReLU activation \cite{Nair:2010:RLU:3104322.3104425} and a layer normalization (LN) \cite{Ba2016Layer}. With the output feature of TTM, the initial prediction block of PPM predicts the immediately following clip feature and action probabilities. The progressive prediction block accumulates the former predictions and the output of TTM, and further predicts a few subsequent future features and actions. The whole TTPP model can be jointly trained in an end-to-end manner with supervision from ground-truth future features and action labels. Compared to previous encoder-decoder methods, the benefits of our TTPP are two-fold. First, the temporal Transformer is more efficient than recurrent methods in capturing historical context by self-attention. Second, the progressive prediction module with skip connections to aggregated historical features can efficiently deliver temporal information and help long-term anticipation.
We evaluate our approach on three widely-used action anticipation datasets, namely TVSeries \cite{De2016Online}, THUMOS-14 \cite{THUMOS14}, and TV-Human-Interaction \cite{doi:10.5244/C.24.50:abbreviated}. Additionally we also conduct a comprehensive study for several popular aggregation and prediction strategies, including temporal convolution, LSTM and single-shot prediction, etc. Extensive results show that TTPP is more efficient than the state-of-the-art methods in both training and inference, and outperforms them with a large margin.

The main contributions of this work can be concluded as follows.
\begin{itemize}
\item We propose a simple yet efficient TTPP framework for action anticipation, which leverages a Transformer-style architecture to aggregate  information and a light-weight module to predict future actions.
\item We elaborately design a progressive prediction module for predicting future features and actions, and achieve the state-of-the-art performance on TVSeries, THUMOS-14, and TV-Human-Interaction.
\item We conduct a comprehensive study for several popular aggregation and prediction strategies, including aggregation methods of temporal convolution, Encoder-LSTM, and prediction methods of Decoder-LSTM, single-shot prediction, etc.
\end{itemize}

The rest of this paper is organized as follows: We first review some related works in Section II. Section III describes the proposed framework with TTM to aggregate observed features and PPM to progressively predict future actions. Afterwards, we show our experimental results on several datasets in section IV and conclude the paper in Section V.

\section{Related Work}
\textbf{Action recognition}. Action recognition is an important branch of video related research areas and has been extensively studied in the past decades. The existing methods are mainly developed for extracting discriminative action features from temporally complete action videos. These methods can be roughly categorized into hand-crafted feature based approaches and deep learning based approaches. Early methods such as Improved Dense Trajectory (IDT) mainly adopt hand-crafted features, such as HOF \cite{Laptev2008Learning}, HOG \cite{Laptev2008Learning} and MBH \cite{Wang2013Dense}. Recent studies demonstrate that action features can be learned by deep learning methods such as convolutional neural networks (CNN) and recurrent neural networks (RNN). Two-stream network \cite{Simonyan2014Two, Wang2015Towards} learns appearance and motion features based on RGB frame and optical flow field separately. RNNs, such as long short-term memory (LSTM) \cite{Graves1997Long} and gated recurrent unit (GRU) \cite{DBLP:journals/corr/ChungGCB14}, have been used to model long-term temporal correlations and motion information in videos, and generate video representation for action classification.  A CNN+LSTM model, which uses a CNN to extract frame features and a LSTM to integrate features over time, is also used to recognize activities in videos\cite{DBLP:journals/corr/DonahueHGRVSD14}. C3D \cite{Du2015Learning} network simultaneously captures appearance and motion features using a series of 3D convolutional layers. Recently, I3D \cite{DBLP:journals/corr/CarreiraZ17} networks use two stream CNNs with inflated 3D convolutions on both dense RGB and optical flow sequences to achieve state of the art performance on Kinetics dataset \cite{DBLP:journals/corr/KayCSZHVVGBNSZ17}.

\textbf{Action anticipation}. Many works have been proposed to exploit the partially observed videos for early action prediction or future action anticipation. Recently, Hoai \etal \cite{Hoai2012Max} propose a max-margin framework with structured SVMs to solve this problem. Ryoo \etal \cite{Ryoo2012Human} develop an early action prediction system by observing some evidences from the temporal accumulated features. Lan \etal \cite{10.1007/978-3-319-10578-9_45} design a coarse-to-ﬁne hierarchical representation to capture the discriminative human movement at different levels, and use a max-margin framework for final prediction. Cao \etal \cite{Yu2013Recognize} formulate the action prediction problem into a probabilistic framework, which aims to maximize the posterior of activity given observed frames. In their work, the likelihood is computed by feature reconstruction error using sparse coding. However, it suffers from high computational complexity as the inference is performed on the entire training data.
Carl \etal \cite{article} present a framework that uses large-scale unlabeled data to predict a rich visual representation in the future, and apply it towards anticipating both actions and objects. Kong \etal \cite{kong2018action} propose a combined CNN and LSTM along with a memory module in order to record ``hard-to-predict'' samples, they benchmark their results on UCF101 \cite{Soomro2012UCF101} and Sports-1M \cite{Karpathy2014Large} datasets. Gao \etal \cite{DBLP:conf/bmvc/GaoYN17a} propose a Reinforced Encoder-Decoder (RED) network for action anticipation, which uses reinforcement learning to encourage the model to make the correct anticipations as early as possible. Ke \etal \cite{Ke_2019_CVPR} propose an attended temporal feature, which uses multi-scale temporal convolutions to process the time-conditioned observation. In this work, we focus only on recent results on anticipation of action labels, more details can be found in \cite{DBLP:conf/bmvc/GaoYN17a} and \cite{Xu_2019_ICCV}.

\begin{figure*}
\begin{center}
\includegraphics[width=1\linewidth]{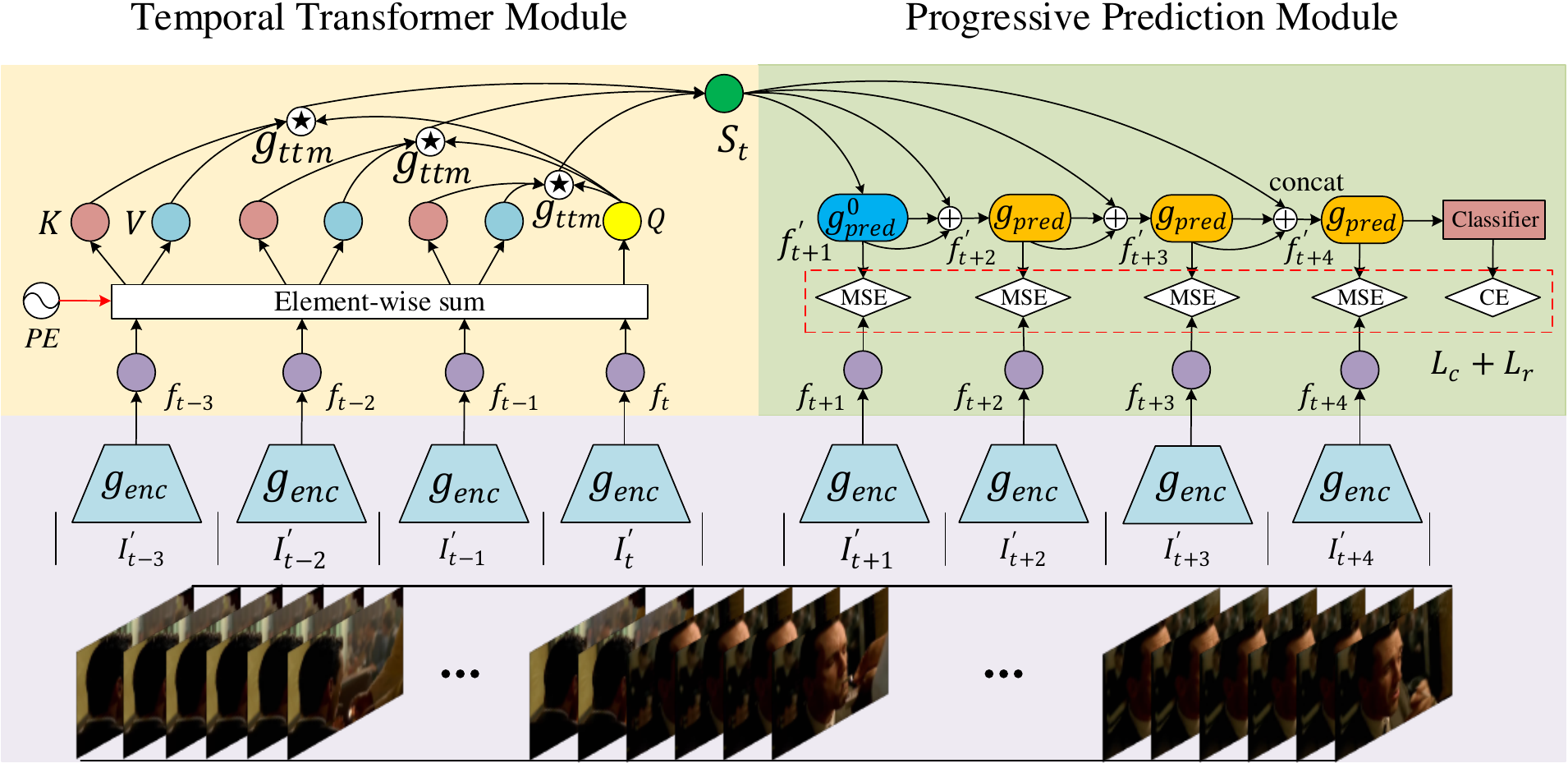}
\end{center}
   \caption{The flowchart of our TTPP method for action anticipation. Given a continuous video sequence, an encoder network is first used to map each video clip to clip features, and then a Temporal Transformer Module is proposed to aggregate observed clip features, and finally a Progressive Prediction Module is designed for future action anticipation. Note that the future information in the red dashed box is only used in training stage and the classifier is performed on each prediction.}
\label{fig_2}
\end{figure*}
\textbf{Online action detection}. Online action detection is usually solved as an online per-frame labelling task on streaming videos, which requires correctly classifying every frame without accessing future frames. De Geest \etal \cite{De2016Online} first introduce the problem by introducing a realistic dataset, \ie TVSeries, and benchmarked the existing models. They have shown that a simple LSTM approach is not sufficient for online action detection, and even worse than the traditional pipeline of improved trajectories, Fisher vectors and SVM. Their later work \cite{De2018} introduces a two-stream feedback network, where one stream processes the input and the other one models the temporal relations. Gao \etal \cite{DBLP:conf/bmvc/GaoYN17a} propose a Reinforced Encoder-Decoder network for action anticipation and treat online action detection as a special case of their framework. Xu \etal \cite{Xu_2019_ICCV} propose the Temporal Recurrent Network (TRN) to model the temporal context by simultaneously performing online action detection and anticipation. Besides, Shou \etal \cite{Shou2018Online} address the online detection of action start (ODAS) by encouraging a classification network to learn the representation of action start windows.

\textbf{Attention for video understanding}.
The attention mechanism which directly models long-term interactions with self-attention has led to state-of-the-art models for action understanding tasks, such as video-based and skeleton-based action recognition \cite{Girdhar_2019_CVPR, Liu2018Skeleton, Liu2017Global, Sharma2017Action, Song2016An}. Our work is related to the recent Video Action Transformer Network \cite{Girdhar_2019_CVPR}, which uses the Transformer architecture as the ``head'' of a detection framework. Specifically, it uses the ROI-pooled I3D feature of a target region as query and aggregates contextual information from the spatial-temporal feature maps of an input video clip. Our work differs from it in the following aspects: (1) The problem is different from spatial-temporal action detection. To the best of our knowledge, we are the first to use Transformer architecture for action anticipation. (2) We have task-specific considerations. For instance, our Transformer unit takes the current frame feature as query and the historical frame features as memory. (3) We elaborately design a light-weight progressive prediction module for efficient action anticipation.
\section{Our Approach}
In this section, we present our temporal Transformer with progressive prediction for the action anticipation task. We propose two module, temporal Transformer module (TTM) to aggregate observed information and progressive prediction module (PPM) to anticipate future actions.
\subsection{Problem Formulation}
The action anticipation task aims to predict the action class $y$ for each frame in the future from an observed action video $V$. More formally, let $V_{1}^{L}=\left [I_{1}, I_{2}, ... , I_{L} \right ]$ be a video with $L$ frames. Given the first $t$ frames $V_{1}^{t}=\left[ I_{1}, I_{2}, ... , I_{t} \right ]$, the task is to predict the actions happening from frames $t+1$ to $L$. That is, we aim to assign action labels $y_{t+1}^{L}=\left[ y_{t+1}, y_{t+2}, ... , y_{L}\right ]$ to each of the unobserved frames.
\subsection{Overall Framework}
Two crucial issues of action anticipation are i) how to aggregate observed information and ii) how to predict future actions. We address these two issues with a simple yet efficient framework, termed as \textit{Temporal Transformer with Progressive Prediction} (TTPP). As illustrated in Figure \ref{fig_2}, a long video is first segmented into multiple non-overlapped chunks $\left[ I_{1}^{'}, I_{2}^{'}, ... , I_{t}^{'}\right]$ with each clip containing an equal number of consecutive frames. Then, a network, \ie $g_{enc}$, maps each video chunk into a representation $f_{t}=g_{enc}(I_{t}^{'})$. More details on video pre-processing and feature extraction are presented in Section \ref{sec_details}. Subsequently, a Temporal Transformer Module (TTM), \ie $g_{ttm}$, temporally aggregates $t$ consecutive chunk representations into a historical  representation $S_{t}=g_{ttm}(f_{1}, f_{2}, ... , f_{t})$. Finally, a Progressive Prediction Module (PPM) progressively predicts future features and actions. The PPM is comprised of an \textit{initial prediction block}, \ie $g_{pred}^{0}$, and a \textit{shared-parameter progressive prediction block}, \ie $g_{pred}$. $g_{pred}^{0}$ takes $S_{t}$ as input and predicts the immediately following clip feature and action probability. $g_{pred}$ accumulates the former predictions and $S_{t}$, and further predicts a few subsequent future features and actions.
\subsection{Temporal Transformer Module (TTM)}
\textbf{Transformer revisit}. Transformer was originally proposed to replace traditional recurrent models for machine translation~\cite{DBLP:journals/corr/VaswaniSPUJGKP17}. The core idea of Transformer is to model correlation between contextual signals by an attention mechanism. Specifically, it aims to encode the input sequence to a higher-level representation by modeling the relationship between queries ($Q$) and memory (keys ($K$) and values ($V$)) with,
\begin{equation}
\label{lable:transformer}
    Attention(Q, K, V)=\textrm{Sofmax}(\frac{QK^{T}}{\sqrt{d_{m}}})V,
\end{equation}
where $Q\in R^{L_q\times d_{m}}$, $K\in R^{L_k\times d_{m}}$ and $V\in R^{L_k\times d_{v}}$. This architecture becomes ``self-attention'' with $Q=K=V=\{f_1,f_2,\cdots, f_T\}$ which is also known as the non-local networks~\cite{Wang2017Non}. A self-attention module maps the sequence to a higher-level representation like RNNs but without recurrence.

\textbf{Temporal Transformer}.To efficiently aggregate observed information, our TTPP framework resorts to a Transformer-style architecture, termed as Temporal Transformer Module (TTM). The TTM takes as input the video chunk features and maps them into a query feature and memory features. For online action anticipation, considering that the last observed feature $f_t$ would be the most relevant one to the future actions, we use $f_t$ as the query of TTM. The memory of TTM is intuitively set as the historical features $[f_{1}, f_{2}, ..., f_{t-1}]$. Formally, the query and memory are as follows,
\begin{equation}
    Q=f_{t},\\
    K=V=[f_{1}, f_{2}, ..., f_{t-1}].
\end{equation}

Since temporal information is lost in the attention operation, we add the positional encoding \cite{DBLP:journals/corr/VaswaniSPUJGKP17} into the input representations. Given sequence feature $f_{in}=\left [f_{1}, f_{2}, ... , f_{T}\right ]\in R^{T\times d_{m}}$, the $i$-th value of the positional vector in temporal position $pos$ is defined as,
\begin{equation}
    PE_{(pos, i)}=\left\{\begin{matrix}
\textrm{sin}(pos/10000^{i/d_{m}}) & \textrm{ if} ~ i ~ \textrm{is even}\\
 \textrm{cos}(pos/10000^{i/d_{m}})& \textrm{ otherwise}.
\end{matrix}\right.
\end{equation}
The original feature vector $f_{pos}$ is then updated by $f_{pos} = f_{pos}+PE_{(pos, :)}$ which provides information about temporal position of each clip feature.

To model complicated action videos, our TTM further leverages the multi-head attention mechanism as follows,
\begin{equation}
\begin{array}{c}
A_t=MultiHead(Q, K, V)=\textrm{Concat}(h_{1}, ..., h_{n})W^{o}, \\
\textrm{where} ~ h_{i}=Attention(QW^{Q}_{i}, KW^{K}_{i}, VW^{V}_{i}),
\end{array}
\end{equation}
where $n$ is the number of attention heads, and $W^{Q}_{i}\in R^{d_{m}\times d_{q}}$, $W^{K}_{i}\in R^{d_{m}\times d_{k}}$, $W^{V}_{i} \in R^{d_{m}\times d_{v}}$ are parameters for the $i$-th attention head which are used for linear projection, and $W^{o}\in R^{nd_{v}\times d_{m}}$ is the projection matrix to reduce the dimension of the concatenated attention vector. For each head, we use $d_{k}=d_{q}=d_{v}=\frac{d_{m}}{n}$. Considering the importance of $f_t$ for anticipation, we view $A_{t}$ as an extra information and add it to the original query feature via a shortcut connection. The final output feature of TTM is $S_{t}=A_{t}+f_{t}$ with dimension $d_{m}$.

\begin{figure}[t]
\begin{center}
\includegraphics[width=9cm]{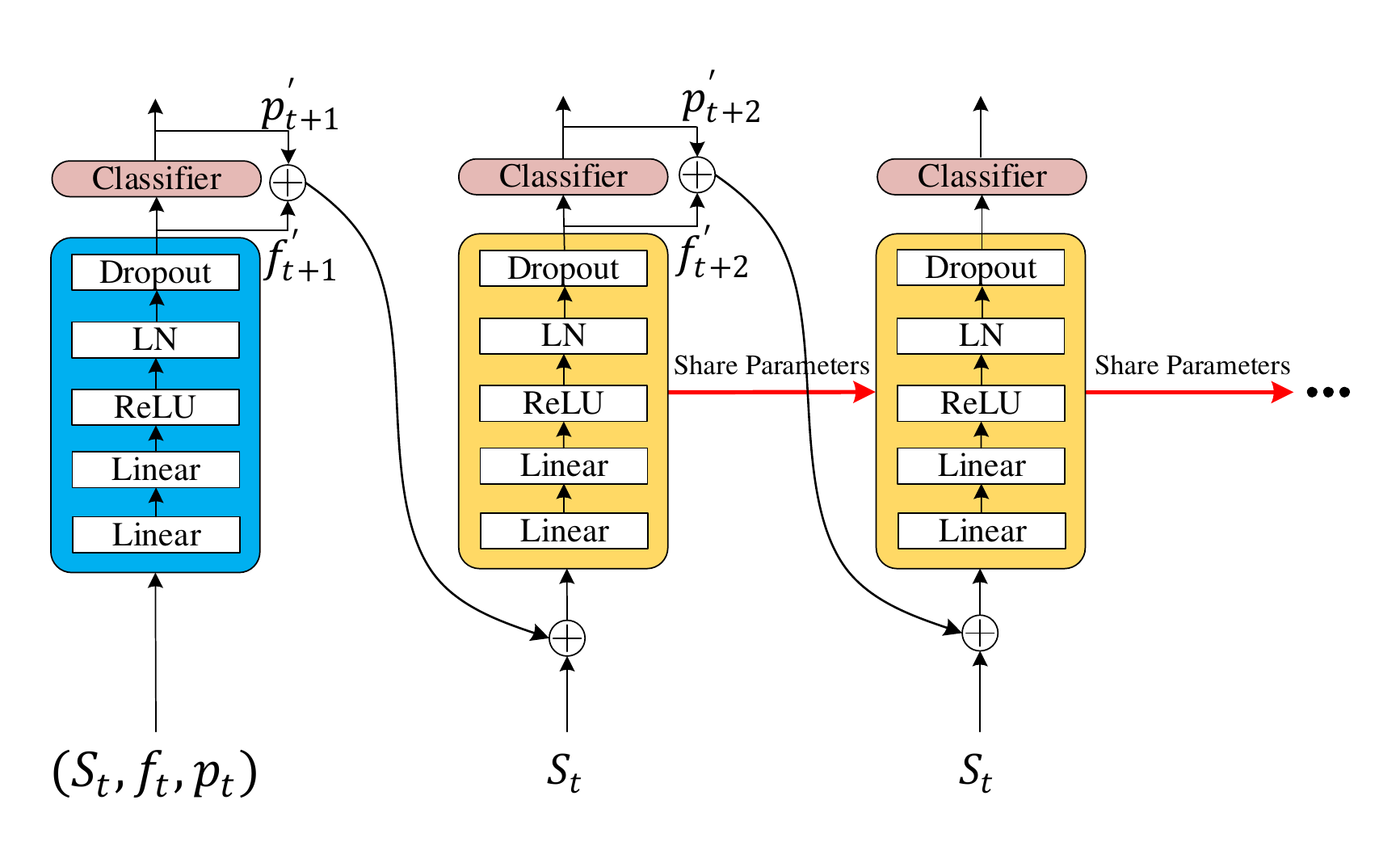}
\end{center}
   \caption{The illustration of our PPM. It consists of an \textit{initial prediction block} highlighted in blue and a \textit{shared-parameter progressive prediction block} highlighted in yellow, where each block is built with two fully-connected layers, followed by a ReLU activation. In addition, we use layer normalization and dropout to improve regularization.}
\label{fig_3}
\end{figure}

\subsection{Progressive Prediction Module (PPM)}
Partially inspired by WaveNet~\cite{DBLP:journals/corr/OordDZSVGKSK16}, we design a Progressive Prediction Module (PPM) to better exploit the aggregated historical knowledge for future prediction. As illustrated in Figure \ref{fig_3}, the PPM is comprised of an \textit{initial prediction block} and \textit{a shared-parameter progressive prediction block}, where each block is built with two fully-connected (FC) layers, a ReLU activation \cite{Nair:2010:RLU:3104322.3104425} and a layer normalization (LN) \cite{Ba2016Layer}.

Assume we predict $l$ steps in the future from time $t+1$ to $t+l$. At the first time step $t+1$, the \textit{initial prediction block} takes as input the aggregated historical representation $S_{t}\in R^{d_m}$ and predicts the feature $f_{t+1}^{'}\in R^{d_m}$ and action probability $p_{t+1}^{'}\in R^{C}$.
Formally, this block is as follows,
\begin{equation}
    p_{t}=\textrm{Softmax}(W_{c}f_{t})
\end{equation}
\begin{equation}
f_{t+1}^{'}=g_{pred}^0(S_{t}\oplus f_{t}\oplus p_{t}),
\end{equation}
\begin{equation}
    p_{t+1}^{'}=\textrm{Softmax}(W_{c}f_{t+1}^{'}),
\end{equation}
where $W_{c}$ is the multi-class ($C$ action classes) action classifier.
At other time step $t+i$ ($i>1$), the previously predicted embedding $f_{t+i-1}^{'}$ and action probability $p_{t+i-1}^{'}$ are first concatenated with $S_{t}$ in channel-wise, and then fed into the \textit{progressive prediction block}. Formally, this block is defined as follows,
\begin{equation}
f_{t+i}^{'}=g_{pred}(S_{t}\oplus f_{t+i-1}^{'}\oplus p_{t+i-1}^{'}),
\end{equation}
\begin{equation}
    p_{t+i}^{'}=\textrm{Softmax}(W_{c}f_{t+i}^{'}),
\end{equation}
where `$\oplus$' denotes concatenate operation. Due to the concatenation, the input dimension of the \textit{progressive prediction block} is $2d_{m}+C$. For both blocks, we use two fully-connected (FC) layers with the first FC reducing the input dimension to $\frac{d_{m}}{2}$ and the second FC generating output vector of dimension $d_m$. It is worth noting that different steps in the \textit{progressive prediction block} share parameters. Thus, the whole PPM is a light-weight network.

\textbf{Training.}
Our TTPP framework is trained in an end-to-end manner with supervision on the PPM module. Specifically, we use two types of loss functions, namely a feature reconstruction loss $L_r$ and a classification loss $L_c$. $L_r$ is the mean squared error loss (MSE) between predicted features and ground-truth features, which is defined as,
\begin{equation}
    L_{r}=\sum_{i=1}^{l}||f_{t+i}^{'}-f_{t+i}||^{2}.
\end{equation}
$L_c$ is the sum of cross-entropy loss (CE) on all the prediction steps, which is defined as,
\begin{equation}
    L_{c}=-\sum_{i=1}^{l}\sum _{j=1}^{C}y_{(t+i, j)}\log p_{(t+i, j)}^{'} ,
\end{equation}
where $y_{(t+i, :)}$ is the one-hot ground-truth vector at time $t+i$.
The total loss is formulated as,
\begin{equation}
    L=L_{c}+\lambda L_{r},
\end{equation}
where $\lambda$ is a trade-off weight for feature reconstruction loss. We experimentally find the final performance is not sensitive to the value of weight, we set $\lambda=1$ for simplicity in our experiments.

\section{Experiments}
The proposed method was evaluated on three datasets, \ie TVSeries \cite{Xu_2019_ICCV}, THUMOS-14 \cite{THUMOS14} and TV-Human-Interaction \cite{doi:10.5244/C.24.50:abbreviated}. We choose these datasets because they include videos from diverse perspectives and applications: TVSeries was recorded from television and contains a variety of everyday activities, THUMOS-14 is a popular dataset of sports-related actions, and TV-Huamn-Interaction contains human interaction actions collected from tv shows.
In this section, we report experimental results and detailed analysis.
\subsection{Datasets}
\textbf{TVSeries} \cite{De2016Online} is originally proposed for online action detection, which consists of $27$ episodes of $6$ popular TV series, namely \textit{Breaking Bad} ($3$ episodes), \textit{How I Met Your Mother} ($8$), \textit{Mad Men} ($3$), \textit{Modern Family} ($6$), \textit{Sons of Anarchy} ($3$), and \textit{Twenty-four} ($4$). It contains totally $16$ hours of video. The dataset is temporally annotated at the frame level with $30$ realistic, everyday actions (\eg, \textit{pick up}, \textit{open door}, \textit{drink}, etc.). It is challenging with diverse actions, multiple actors, unconstrained viewpoints, heavy occlusions, and a large proportion of non-action frames.

\textbf{THUMOS-14} \cite{THUMOS14} is a popular benchmark for temporal action detection. It contains over $20$ hours of sport videos annotated with $20$ actions. The training set (\ie UCF101 \cite{Soomro2012UCF101}) contains only trimmed videos that cannot be used to train temporal action detection models. Following prior works \cite{DBLP:conf/bmvc/GaoYN17a, Xu_2019_ICCV}, we train our model on the validation set (including $3K$ action instances in $200$ untrimmed videos) and evaluate on the test set (including $3.3K$ action instances in $213$ untrimmed videos).

\textbf{TV-Human-Interaction} (TV-HI) \cite{doi:10.5244/C.24.50:abbreviated}.
We also evaluate our method on TV-Human-Interaction which is also used in \cite{DBLP:conf/bmvc/GaoYN17a}. The dataset contains $300$ trimmed video clips extracted from $23$ different TV shows. It is annotated with four interaction classes, namely \textit{hand shake}, \textit{high five}, \textit{hug}, and \textit{kissing}. It also contains a \textit{negative} class with $100$ videos, that have none of the listed interactions. We use the suggested experimental setup of two train/test splits.

\subsection{Evaluation Protocols}
For each class on TVSeries, we use the per-frame calibrated average precision (cAP) which is proposed in \cite{De2016Online},
\begin{equation}
    cAP=\frac{\sum \nolimits_{k}cPrec(k)\ast I(k)}{P},
\end{equation}
where calibrated precision $cPrec=\frac{TP}{TP+FP/w}$, $I(k)$ is an indicator function that is equal to $1$ if the cut-off frame $k$ is a true positive, $P$ denotes the total number of true positives, and $w$ is the ratio between negative and positive frames. The mean cAP over all classes is reported for final performance. The advantage of cAP is that it is fair for class imbalance condition. For THUMOS-14, we report per-frame mean Average Precision (mAP) performance. For TV-Human-Interaction, we report classification accuracy (ACC).

\subsection{Implementation Details}
\label{sec_details}
To make fair comparisons with state-of-the-art methods \cite{De2016Online, DBLP:conf/bmvc/GaoYN17a, Xu_2019_ICCV}, we follow their experimental settings on each dataset.

\textbf{Chunk-level feature extraction}. We extract frames from all videos at $24$ Frames Per Second ($FPS$). The video chunk size is set to $6$, \ie $0.25$ second. We use three different feature extractors as the visual encoder $g_{enc}$, VGG-16  \cite{Simonyan2014Very} network pre-trained on UCF101 \cite{Soomro2012UCF101}, two-stream (TS) \cite{Xiong2016CUHK} network\footnote{https://github.com/yjxiong/anet2016-cuhk} pre-trained on ActivityNet-1.3 \cite{Caba2015ActivityNet}, and inflated 3D ConvNet\footnote{https://github.com/piergiaj/pytorch-i3d} (I3D) \cite{Carreira_2017_CVPR} pre-trained on Kinetics \cite{Kay2017The}. VGG-16 features (4096-D) are extracted at the $fc6$ layer for the central frame of each chunk. For the two-stream features in each chunk, the appearance CNN feature is extracted on the central frame which is the output of \textit{Flatten 673 layer} in ResNet-200 \cite{DBLP:journals/corr/HeZRS15}, and the motion feature is extracted on the $6$ optical flow frames of each chunk which is output of \textit{global pool layer} in a pre-trained BN-Inception model \cite{Ioffe2015Batch}. The motion feature and appearance feature are then concatenated into a TS feature (4096-D) for each chunk. Different from prior works \cite{DBLP:conf/bmvc/GaoYN17a, Xu_2019_ICCV}, we also use recent I3D features. The I3D model is originally trained with $64$-frame video snippets, thus may not be a good idea for per-frame action anticipation. Nevertheless, we input the $6$ frames of each chunk to I3D and extract the output (1024-D) of the last \textit{global average pooling layer} as I3D-based feature.

\textbf{Hyperparameter setting}. We implement our proposed method in PyTorch and perform all experiments on a system with $8$ Nvidia TiTAN X graphic cards. We use the SGD optimizer with a learning rate $0.001$, a momentum of $0.9$, and batch size $32$. The input sequence length is set to $8$ by default, corresponding to $2$ seconds. We use single-layer multi-head setting for our TTM, and the number of heads is set to $4$ by default.

\begin{table}[tp]
\centering
\resizebox{\linewidth}{!}{
\begin{tabular}{@{}lllllllllll@{}}
\toprule
& & \multicolumn{8}{l}{Time predicted into the future (seconds)} &     \\ \cmidrule(l){3-10}
  Method      &      Inputs                   &       0.25s& 0.5s& 0.75s& 1.0s& 1.25s& 1.5s& 1.75s& 2.0s    & Avg \\ \midrule
  ED~\cite{DBLP:conf/bmvc/GaoYN17a}        &     VGG        &      71.0& 70.6& 69.9& 68.8& 68.0& 67.4& 67.0& 66.7& 68.7     \\
  RED~\cite{DBLP:conf/bmvc/GaoYN17a}   &         VGG       &      71.2& 71.0& 70.6& 70.2& 69.2& 68.5& 67.5& 66.8& 69.4     \\
  Ours     &       VGG        &       \textbf{72.7}& \textbf{72.3}& \textbf{71.9}& \textbf{71.6}& \textbf{71.3}& \textbf{70.9}& \textbf{69.9}& \textbf{69.3}& \textbf{71.3 }   \\ \midrule
  ED~\cite{DBLP:conf/bmvc/GaoYN17a}        &     TS        &      78.5& 78.0& 76.3& 74.6& 73.7& 72.7& 71.7& 71.0& 74.5     \\
      RED~\cite{DBLP:conf/bmvc/GaoYN17a}      &     TS         &  79.2& 78.7& 77.1& 75.5& 74.2& 73.0& 72.0& 71.2& 75.1    \\
      TRN \cite{Xu_2019_ICCV}   &    TS               &      79.9& 78.4& 77.1& 75.9& 74.9& 73.9& 73.0& 72.3& 75.7     \\
         Ours               &    TS               &    \textbf{81.2}& \textbf{80.3}& \textbf{79.3}& \textbf{77.6}&\textbf{ 76.9}& \textbf{76.7}& \textbf{76.0}& \textbf{74.9}& \textbf{77.9}    \\ \bottomrule
\end{tabular}}
\caption{Comparison with the state-of-the-art methods on TVSeries in terms of mean cAP (\%).}
\label{tab:tvseries}
\end{table}

\begin{table}[tp]
\centering
\resizebox{\linewidth}{!}{
\begin{tabular}{@{}lllllllllll@{}}
\toprule
& & \multicolumn{8}{l}{Time predicted into the future (seconds)} &     \\ \cmidrule(l){3-10}
  Method      &      Inputs                   &       0.25s& 0.5s& 0.75s& 1.0s& 1.25s& 1.5s& 1.75s& 2.0s    & Avg \\ \midrule
  ED~\cite{DBLP:conf/bmvc/GaoYN17a}        &     TS        &      43.8& 40.9& 38.7& 36.8& 34.6& 33.9& 32.5& 31.6& 36.6     \\
      RED~\cite{DBLP:conf/bmvc/GaoYN17a}      &     TS         &  45.3& 42.1& 39.6& 37.5& 35.8& 34.4& 33.2& 32.1& 37.5    \\
      TRN \cite{Xu_2019_ICCV}   &    TS               &      45.1& 42.4& 40.7& 39.1& 37.7& 36.4& 35.3& 34.3& 38.9     \\
         Ours               &    TS               &    \textbf{45.9}& \textbf{43.7}& \textbf{42.4}& \textbf{41.0}& \textbf{39.9}& \textbf{39.4}& \textbf{37.9}& \textbf{37.3}& \textbf{40.9}    \\ \midrule
         Ours               &    I3D               &    46.8& 45.5& 44.6& 43.6& 41.9& 41.1& 40.4& 38.7& \textbf{42.8}    \\ \bottomrule
\end{tabular}}
\caption{Comparison with the state-of-the-art methods on THUMOS-14 in terms of mAP (\%).}
\label{tab:thumos}
\end{table}

\begin{table}[tp]
\centering
\small
\resizebox{\linewidth}{!}{
\begin{tabular}{ccccc}
\toprule
 Method & Vondrick \etal \cite{article}  & RED-VGG \cite{DBLP:conf/bmvc/GaoYN17a}  & RED-TS \cite{DBLP:conf/bmvc/GaoYN17a}  & Ours-TS \\  \midrule
ACC  (\%) & 43.6 & 47.5 & 50.2  & \textbf{53.5} \\ \bottomrule
\end{tabular}}
\caption{Anticipation results on TV-Human-Interaction at $T_{a}=1s$ in terms of ACC (\%).}
\label{tab:tv}
\end{table}

\subsection{Popular Baselines}
Here we present several advanced baselines for temporal information aggregation and future prediction.

\textbf{Temporal convolution} (\ie Conv1D) aggregates temporal features with 1-D convolution operations in temporal axis. We apply $3$ Conv1D layers with kernel size $3$ and stride $2$ on two-stream features for this baseline.

\textbf{LSTM} takes sequence features as input and recurrently updates its hidden states over time. The \textbf{Encoder-LSTM} summarizes historical information into the final hidden state for information aggregation. The \textbf{Decoder-LSTM} recurrently decodes information into hidden states as predicted features. We use a single-layer LSTM architecture with $4096$ hidden units for this baseline.

\textbf{Single-shot prediction} (SSP). We implement a single-shot prediction
method similar to \cite{DBLP:conf/bmvc/GaoYN17a, article}. With the aggregated historical feature, this method uses two FC layers to anticipate the single future feature at $T_{a}$, where $T_{a}\in \left \{ t+1, t+2, ..., t+l \right \}$. This prediction method is equal to our PPM without the progressive process.


\subsection{Comparison with State of the Art}
We compare our proposed TTPP method to several state-of-the-art methods on TVSeries, THUMOS-14, and TV-HI. The results are presented in Table \ref{tab:tvseries}, Table \ref{tab:thumos}, and Table \ref{tab:tv}, respectively. Our method consistently outperforms all these methods in all the predicted steps. With two-stream features, our TTPP achieves $77.9\%$ (mean cAP), $40.9\%$ (mAP), and  $53.5\%$ (ACC) on these datasets, which outperforms these recent advanced methods by $2.2\%$, $2.0\%$, and $3.3\%$, respectively.

On both TVSeries and THUMOS-14, the improvements over other methods are more evident on long-term predictions. For instance, with two-stream features, our TTPP outperforms ED (Encoder-Decoder LSTM)~\cite{DBLP:conf/bmvc/GaoYN17a} by $2.1\%$ at $T_a=0.25s$ and $5.7\%$ at $T_a=2.0s$ on THUMOS-14, and these numbers are $2.7\%$ and $3.9\%$ on TVSeries. With VGG features, our method improves the Reinforcement ED by $2.6\%$ in average cAP on TVSeries. Since the VGG and TS features are relatively old, we also test the I3D features, which updates a new state-of-the-art on THUMOS-14 with $42.8\%$ in average mAP over time.

\begin{table}[tp]
\centering
\begin{subtable}[t]{1\linewidth}
\centering
\resizebox{\linewidth}{!}{
\begin{tabular}{@{}llllllllll@{}}
\toprule
&  \multicolumn{8}{l}{Time predicted into the future (seconds)} &     \\ \cmidrule(l){2-9}
  Method & 0.25s& 0.5s& 0.75s& 1.0s& 1.25s& 1.5s& 1.75s& 2.0s & Avg \\ \midrule
Conv1D$_{-}$ & 77.9& 77.1& 75.9& 74.3& 73.6& 72.7& 71.9& 71.0& 74.3   \\
Conv1D & 79.4& 77.9& 76.6& 75.4& 73.9& 73.6& 72.8& 72.4& 75.3    \\
LSTM$_{-}$ & 79.3 & 78.2& 76.9& 74.6& 73.0& 72.3& 71.6 & 69.8 & 74.5   \\
LSTM &  78.9& 77.9& 76.3& 75.5& 74.3& 73.8& 72.8& 72.0& 75.2    \\
TTM$_{-}$  &79.1  &78.6 & 77.9& 77.0& 76.4&75.6 & 75.1& 74.1& 76.7\\
TTM (Ours) & 81.2& 80.3& 79.3& 77.6& 76.9& 76.7& 76.0& 74.9& \textbf{77.9}     \\\bottomrule
\end{tabular}}
\caption{TVSeries}
\end{subtable}
\qquad
\begin{subtable}[t]{1\linewidth}
\centering
\resizebox{\linewidth}{!}{
\begin{tabular}{@{}llllllllll@{}}
\toprule
&  \multicolumn{8}{l}{Time predicted into the future (seconds)} &     \\ \cmidrule(l){2-9}
  Method                      &       0.25s& 0.5s& 0.75s& 1.0s& 1.25s& 1.5s& 1.75s& 2.0s    & Avg \\ \midrule
         Conv1D                             &    41.8& 40.9& 39.6& 38.1& 37.2& 36.7& 35.9& 35.5& 38.2    \\
         LSTM                              &    41.6& 40.5& 39.1& 37.9& 35.6& 34.9& 34.3& 33.3& 37.0    \\
         TTM (Ours)                          &    45.9& 43.7& 42.4& 41.0& 39.9& 39.4& 37.9& 37.3 &\textbf{40.9}    \\ \bottomrule
\end{tabular}}
\caption{THUMOS-14}
\end{subtable}
\caption{Evaluation of temporal \textit{aggregation} methods with two-stream features on TVSeries and THUMOS-14. For fair comparison, PPM is used for future prediction. ``$_{-}$'' denotes that the aggregated feature is directly used for prediction without the shortcut connection to current feature.}
\label{tab:aggregation}
\end{table}

\begin{table}[tp]
\centering
\begin{subtable}[t]{1\linewidth}
\centering
\resizebox{\linewidth}{!}{
\begin{tabular}{@{}llllllllll@{}}
\toprule
&  \multicolumn{8}{l}{Time predicted into the future (seconds)} &     \\ \cmidrule(l){2-9}
  Method (A-P)$^*$                         &       0.25s& 0.5s& 0.75s& 1.0s& 1.25s& 1.5s& 1.75s& 2.0s    & Avg \\ \midrule
  LSTM-LSTM (ED~\cite{DBLP:conf/bmvc/GaoYN17a})                &      78.5& 78.0& 76.3& 74.6& 73.7& 72.7& 71.7& 71.0& 74.5    \\
      LSTM-SSP (EFC~\cite{DBLP:conf/bmvc/GaoYN17a})           &  78.4& 76.9& 74.2& 72.7& 70.7& 70.2& 69.0& 67.9& 72.5    \\
      LSTM-PPM                &      78.6& 77.5& 76.0& 75.0& 73.5& 73.8& 72.8& 71.8& 74.9     \\\midrule
         TTM-LSTM                         &  79.3  & 77.5& 76.2& 75.1& 73.3& 72.8& 71.6& 69.9& 74.5    \\
         TTM-SSP                          &    80.1& 77.1& 75.3& 73.6& 72.3& 71.7& 70.0& 68.9& 73.6    \\
         TTM-PPM (Ours)                    &    81.2& 80.3& 79.3& 77.6& 76.9& 76.7& 76.0& 74.9& \textbf{77.9}    \\ \bottomrule
\end{tabular}}
\caption{TVSeries}
\end{subtable}
\\
\begin{subtable}[t]{1\linewidth}
\centering
\resizebox{\linewidth}{!}{
\begin{tabular}{@{}llllllllll@{}}
\toprule
&  \multicolumn{8}{l}{Time predicted into the future (seconds)} &     \\ \cmidrule(l){2-9}
  Method (A-P)$^*$                         &       0.25s& 0.5s& 0.75s& 1.0s& 1.25s& 1.5s& 1.75s& 2.0s    & Avg \\ \midrule
  LSTM-LSTM (ED~\cite{DBLP:conf/bmvc/GaoYN17a})                &      43.8& 40.9& 38.7& 36.8& 34.6& 33.9& 32.5& 31.6& 36.6    \\
      LSTM-SSP (EFC~\cite{DBLP:conf/bmvc/GaoYN17a})           &  40.6& 39.3& 37.2& 34.9& 33.2& 31.5& 30.4& 28.5& 34.4    \\
      LSTM-PPM                 &      41.3& 40.3& 38.9& 37.6& 35.4& 34.6& 33.9& 33.0& 36.8     \\\midrule
         TTM-LSTM                         & 44.4   &43.1& 41.3& 40.2& 38.7& 37.9& 37.2& 36.6 & 39.9  \\
         TTM-SSP                          &    43.9& 41.0& 38.5& 37.1& 35.5& 32.7& 32.2& 30.5& 36.4    \\
         TTM-PPM (Ours)                    &    45.9& 43.7& 42.4& 41.0& 39.9& 39.4& 37.9& 37.3& \textbf{40.9}    \\ \bottomrule
\end{tabular}}
\caption{THUMOS-14}
\end{subtable}
\caption{Evaluation of future \textit{prediction} methods with two-stream features on TVSeries and THUMOS-14. Both TTM and LSTM are evaluated for temporal aggregation. (A-P)$^*$ denotes temporal aggregation and future prediction, respectively.}
\label{tab:prediction}
\end{table}
\subsection{Ablation Study of TTM and PPM}

To further investigate the effectiveness of our proposed TTPP, we conduct extensive evaluations for both TTM and PPM by comparing them to recent temporal aggregation and prediction methods, respectively.

For temporal aggregation, we compare our TTM to Conv1D and Encoder-LSTM on both THUMOS-14 and TVSeries with the PPM as prediction phase. Since we use a shortcut connection in TTM to highlight the current frame information, we also evaluate the benefits of this design for all the aggregation methods. The results are shown in Table \ref{tab:aggregation}. Several observations can be concluded as follows. First, the proposed TTM is superior to both Conv1D and LSTM regardless of the shortcut connection. Specifically, TTM outperforms Conv1D and LSTM by $2.6\%$ ($2.7\%$) and $2.7\%$ ($3.9\%$) with shortcut connection on TVSeries (THUMOS-14), respectively. Second, the shortcut connection to current feature significantly improves all methods on TVSeries. For instance, our TTM degrades from $77.9\%$ to $76.7\%$ after removing the shortcut connection which demonstrates the importance of current feature and the superiority of our design. Last but not the least, the improvements of our TTM over other methods are similar for different time steps, which suggests that TTM provides better aggregated features via attention than the others.

For future prediction, we compare our PPM to Decoder-LSTM and SSP with either Encoder-LSTM or TTM as aggregation method. The results are presented in Table \ref{tab:prediction}. Several finds are concluded as follows. First, with both aggregation methods, our PPM consistently outperforms Decoder-LSTM and SSP on both datasets which shows the effectiveness of PPM. Second, our PPM obtains more improvements with our aggregation method TTM than Encoder-LSTM. For instance, TTM-PPM outperforms TTM-LSTM by $3.4\%$ while LSTM-PPM only outperforms LSTM-LSTM by $0.4\%$. Third, with both aggregation methods, our PPM is significantly superior to SSP on both datasets especially at long-term prediction time steps, which demonstrates the progressive design of our PPM is important.

\begin{figure}
\begin{center}
\includegraphics[width=0.7\linewidth]{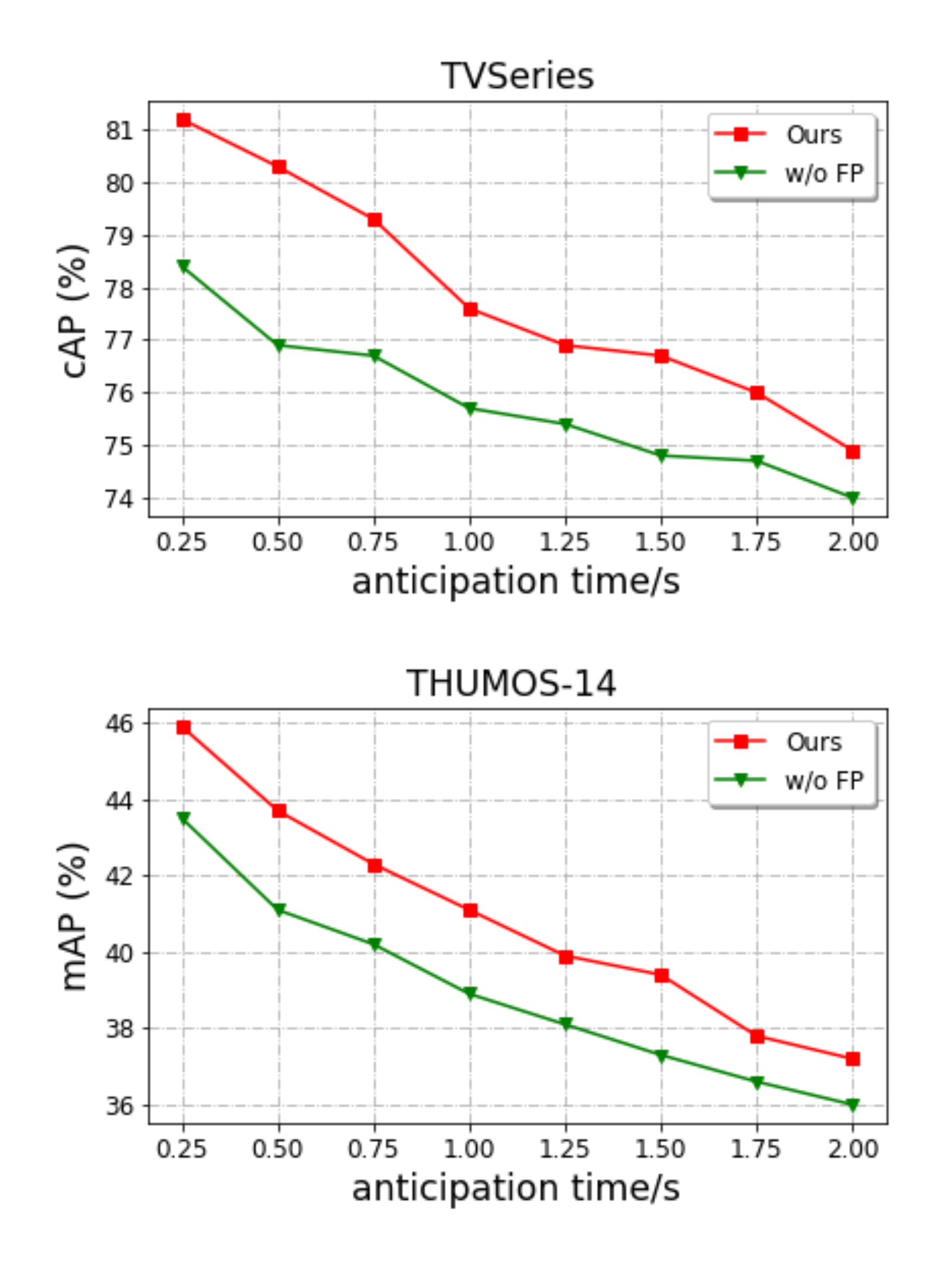}
\end{center}
  \caption{Evaluation of feature prediction for action anticipation on TVSeries (cAP \%) and THUMOS-14 dataset (mAP \%) with two-stream features.}
\label{fig_FP}
\end{figure}

\subsection{Importance of Feature Prediction}
In order to evaluate the influence of feature prediction for the final action anticipation, we remove the predicted features (w/o FP) by only use the concatenation of action probability and the aggregated historical representation in the PPM. The results are shown in Figure \ref{fig_FP}. Without considering the predicted features, the performance of the model w/o FP degenerates dramatically. It indicates that only relying on the action probability to predict future actions is not enough and the predicted feature representations are always related to the action itself and thus could possibly provide some useful information.

\begin{figure}
\begin{center}
\includegraphics[width=0.7\linewidth]{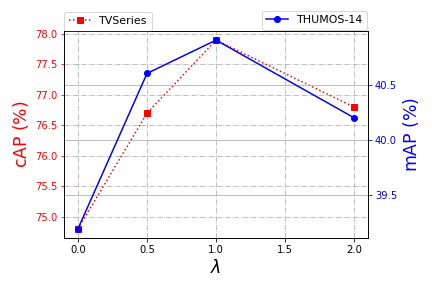}
\end{center}
  \caption{Evaluation of trade-off weight $\lambda$ on TVSeries (cAP \%) and THUMOS-14 dataset (mAP \%) with two-stream features.}
\label{fig_lambda}
\end{figure}

\begin{figure}
\begin{center}
\includegraphics[width=0.7\linewidth]{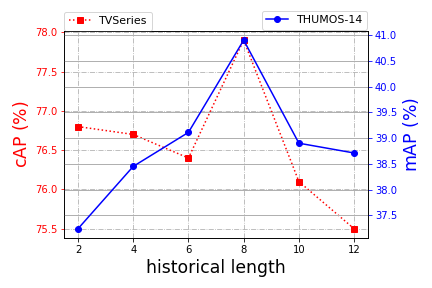}
\end{center}
  \caption{Evaluation of input sequence length for temporal aggregation on TVSeries (cAP \%) and THUMOS-14 dataset (mAP \%) with two-stream features.}
\label{fig_length}
\end{figure}

\begin{figure*}
\begin{center}
\includegraphics[width=1.0\linewidth]{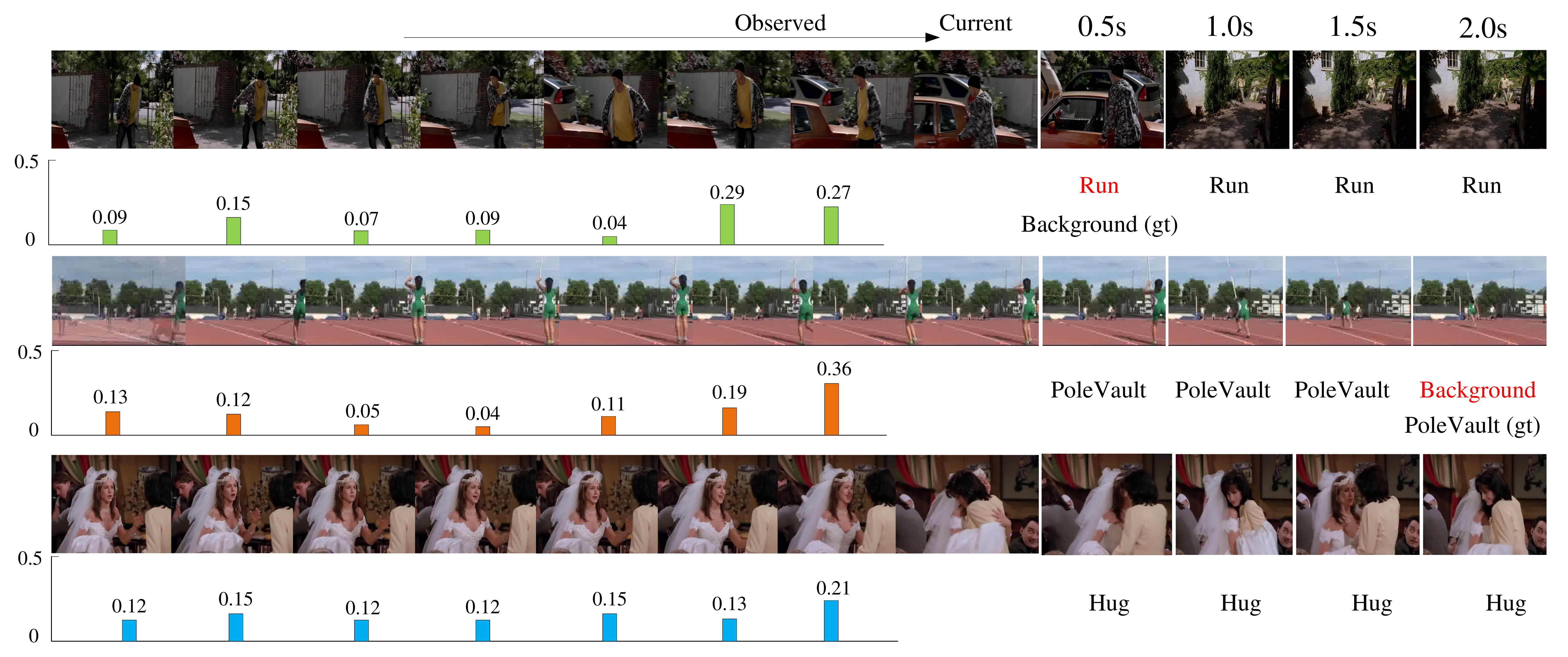}
\end{center}
  \caption{Visualization of attention weights and action anticipation on TVSeries (1st row), THUMOS-14 (2nd row), and TV-Human-Interaction (3rd row). Incorrect anticipation results are shown in red.}
\label{fig_4}
\end{figure*}

\subsection{Evaluation of Sequence Length and Parameters}
\label{sec_length}
In the above experiments, we use a fixed historical length $8$ for aggregation, $4$ parallel heads and trade-off loss weight $\lambda =1.0$ for training by default. To investigate their impacts to the proposed TTPP framework, we evaluate them on both THUMOS-14 and TVSeries.

\textbf{The impact of $\lambda$}.
$\lambda$ is the weight of feature reconstruction loss in training. Figure \ref{fig_lambda} shows the results of varied $\lambda$ on THUMOS-14 and TVSeries.
Removing the feature reconstruction loss, \ie $\lambda=0$, degrades performance dramatically on both datasets which suggests the necessary of feature prediction.
Increasing the weight from 0 to 1 improves performance, and it gets saturation or slightly hurts performance after 1. This may be explained by that overemphasizing feature reconstruction can hurt the discrimination of predicted features.

\textbf{Number of heads $n$}. We also study performance variations given various number of heads for temporal Transformer. Average prediction performance of our TTPP network with $n\in\{1,2,4,8,16\}$ are shown in Table \ref{tab_head}. Results in Table \ref{tab_head} indicate that our method is not sensitive to parameter $n$. The largest performance variation is only within $0.8\%$ on TVSeries and $1.1\%$ on THUMOS-14. On both datasets, we achieve best performance with head number $n=4$.

\textbf{Input sequence length}.
The length of observed sequence determines how much historical information can be used. Figure \ref{fig_length} illustrates the evaluation results on THUMOS-14 and TVSeries. On both datasets, we achieve the best performance with length 8. Decreasing sequence length leads to insufficient context information and increasing sequence length results to massive background information which are both inferior to the default length.

\subsection{Efficiency and Visualization}
Table \ref{tab_5} reports a comparison of parameters, memory footprint, inference time and performance of different models on TVSeries dataset. Compared to the popular Encoder-Decoder LSTM model, our TTPP has $64\%$ fewer parameters, $44\%$ fewer memory footprint and less inference time, while achieves $4.6\%$ higher performance. The efficiency of the proposed TTPP owes to both the Transformer architecture for sequence modeling and the efficient progressive prediction module.

Figure \ref{fig_4} shows some examples of attention weights and action anticipation on TVSeries, THUMOS-14 and TV-Human-Interaction. We find that frames near the current frame usually get higher weights compared to these distant frames since the current frame feature is used as the query.
On TVSeries and THUMOS-14, multiple action instances and confusing background frames exist in the videos which lead to incorrect anticipation inevitably.

\begin{table}[tp]
\centering
\resizebox{\linewidth}{!}{
\begin{tabular}{cccccc}
\toprule
Number of Heads & n=1 & n=2 & n=4 & n=8 & n=16 \\ \midrule
TVSeries & 77.1 & 77.7 & \textbf{77.9} & 77.5 & 77.2 \\
THUMOS-14  & 40.1  & 40.4 & \textbf{40.9} & 40.2 & 39.8 \\ \bottomrule
\end{tabular}}
\caption{Comparison between different number of heads on TVSeries and THUMOS-14 with two-stream features.}
\label{tab_head}
\end{table}

\begin{table}[tp]
\centering
\resizebox{\linewidth}{!}{
\begin{tabular}{ccccc}
\toprule
Model & Parameter (M) & Memory (M) & Inf (s) & \multicolumn{1}{l}{cAP (\%)} \\ \midrule
ED \cite{DBLP:conf/bmvc/GaoYN17a} & 277 & 6560 & 212 & 74.5 \\
TTPP & 101 & 3675 & 145 & 77.9 \\ \bottomrule
\end{tabular}}
\caption{Comparison of parameter, memory footprint and inference time on TVSeries dataset with two-stream features.}
\label{tab_5}
\end{table}

\section{Conclusion}
In this paper, we propose a novel deep framework to boost action anticipation by adopting Temporal Transformer with Progressive Prediction, where a TTM is used to aggregate observed information and a PPM to progressively predict future features and actions. Experimental results on TVSeries, THUMOS-14, and TV-Human-Interaction demonstrate that our framework significantly outperforms the state-of-the-art methods. Extensive ablation studies are conducted to show the effectiveness of each module of our method.

\section{Acknowledgements}
This work is partially supported by the National Key Research and Development Program of China (No.2016YFC1400704), and National Natural Science Foundation of China (U1813218, U1713208, 61671125), Shenzhen Basic Research Program (JCYJ20170818164704758, CXB201104220032A), the Joint Lab of CAS-HK, Shenzhen Institute of Artificial Intelligence and Robotics for Society, and the Sichuan Province Key Rresearch and Development Plan (2019YFS0427).

\bibliography{refs}
\end{document}